\documentclass{article}

\usepackage{arxiv}
\usepackage{graphicx} 
\usepackage{subfigure} 
\usepackage{amsfonts}
\usepackage{url}

\usepackage{algorithm}
\usepackage{algorithmic}
\usepackage{multirow}
\usepackage{tikz}
\usepackage{amsmath}
\usepackage{amsthm}
\usetikzlibrary{bayesnet}
\usetikzlibrary{arrows,topaths}
\usepackage{verbatim}  

\usepackage{hyperref}

\usepackage[utf8]{inputenc} 
\usepackage[T1]{fontenc}    
\usepackage{hyperref}       
\usepackage{url}            
\usepackage{booktabs}       
\usepackage{amsfonts}       
\usepackage{nicefrac}       
\usepackage{microtype}      
\usepackage{lipsum}
\usepackage{amsmath}
\title{Learning Representations of Graph Data: \\A Survey}

\author{
 Mital Kinderkhedia \\
  Department of Statistical Science\\
  University College London\\
  London, W1CE 6BT \\
  \texttt{mital.kinderkhedia.10@ucl.ac.uk} \\
}

\begin{document}
\maketitle

\begin{abstract}
Deep Neural Networks have shown tremendous success in the area of object recognition, image classification and natural language processing. However, designing optimal Neural Network architectures that can learn and output arbitrary graphs is an ongoing research problem. The objective of this survey is to summarize and discuss the latest advances in methods to Learn Representations of Graph Data. We start by identifying commonly used types of graph data and review basics of graph theory. This is followed by a discussion of the relationships between graph kernel methods and neural networks. Next we identify the major approaches used for learning representations of graph data namely: Kernel approaches, Convolutional approaches, Graph neural networks approaches, Graph embedding approaches  and Probabilistic approaches. A variety of methods under each of the approaches are discussed and the survey is concluded with a brief discussion of the future of learning representation of graph data.
\end{abstract}


\section{Introduction}
Structing data into a graph facilitates the study of uncovering complex relationships  and patterns in a systematic manner. For example, a World Wide Web graph shows complex structures given the high frequency of the links between webpages \cite{web} and in Natural Language Processing \cite{tree} text is sometimes represented using trees to understand linkages between words to infer meaning of sentences. However, research in machine learning primarily focuses on data represented in a vectorial form. Real world data is not easily represented as vectors. Examples of real world scenarios with complex graph like structures include molecular and biological networks, computer networks, sensor networks, social networks, citation networks, power grids and transportation networks. Using a graph-based representation, it is possible to capture the sequential, topological, geometric and other relational characteristics of structured data. 
\\
\\
\emph{Neural Networks} are universal function approximators \cite{uni}. Following recent advances, deep learning models have achieved tremendous success in speech recognition \cite{dhal2},\cite{dhal1},\cite{deng} object recognition-detection \cite{housenumbers},\cite{imagenet}, and in learning natural language processing. Further, a  confluence of ingredients: large datasets, advanced computational processing power and burgeoning research in machine learning methods has tremendously contributed to deep learning research. The main distinction we point out between neural and non-neural methods for machine learning is in the \emph{learning representations of data} \cite{bengio}.  In machine learning terminology, we use the term features, whereas in representation learning terminology, we are concerned with learning the most optimal  representations of data which make downstream machine learning tasks proficient. 
The idea behind learning \emph{graph representations} is to learn a type of mapping, one such that embeds the vertices, subgraphs or whole graphs into points in a low-dimensional vector space. These mappings are then optimised so that they reflect the geometric structure within the embedding space and the learned embeddings can then be used as vectorial inputs for machine learning tasks. 
 
\subsection{Contribution}
The contribution of this survey is to propose a taxonomy of major approaches to learn graph representations identified as follows: Kernel approaches, Convolution approaches, Graph neural network approaches, Graph embedding approaches and Probabilistic approaches. This paper compares, contrasts and outlines the techniques used in these approaches. However, this survey is not exhaustive.

\subsection{GRAPH DATA DOMAINS}
Popular data domains that use graph-based representations are discussed, though the list is not exhaustive.
\\
\\
\textbf{Biological Data}:
Typical biological data represents sequences of DNA, sequences of RNA and tertiary structure of proteins.  The motivation for analysing such data is to discover new biological insights. Biological networks, such as the metabolic network \cite{motifgraphs} of bacteria \emph{Escherichia coli} can be modelled as a graph to learn the relationships between small biomolecules (metabolites) and enzymes (proteins). Given that protein-protein interaction is crucial for majority of biological processes, Yook et al study \emph{Saccharomyces cerevisiae} with the aim to uncover the
network`s generic large-scale properties and the impact of the protein`s function and cellular
localisation on the network topology \cite{oltvai}.
  \\
  \\
\textbf{Chemical Data}:
To represent chemical compounds as graph structures, the vertices can represent the atoms and the edges correspond to the bonds. In  \cite{nale}, graphs are used to model the the key topological and
geometric characteristics of chemical structures. The motivation for  modelling a set of chemical compounds or molecules as graphs is to understand  their key characteristics, their toxicity and biological activity, for example. This has become a focal-point area of chemical graph mining \cite{tox}. Chemical data are unique in structure and the typical applications include mining substructures for the comparison of chemical compounds, predicting compound bio-activity using classification, regression and ranking \cite{chemo}.
\\
\\
\textbf{Web Data}:
Typical web data are in the form of hypertext documents, shopping histories, browsing histories and search histories, for example. A webpage, is represented as a vertex and the hyperlinks between the pages as edges. Web pages modelled as graph objects are for the purpose of capturing the linkage structure. Given vast amounts of internet data, one goal would be to develop models that leverage the network topology in order to extract relational knowledge \cite{zanghi}.
\\
\\
\textbf{Text Data}:
Unprecedented growth of text data on the web has given room to model such data in a variety of ways. One approach is to use probabilistic (or neural probabilistic) models. Another approach is to use graphs. For example, in one case words could be represented as vertices and relations between the words as edges, and in another, topics represented as vertices and edges as relations between those topics. Vertices can also denote features of text. Examples of some popular graphs using text data include the co-occurrence graph, the semantic graph and the hierarchical keyword graph \cite{graphtext}. Organisations that leverage machine learning models, to learn from text, stand to gain insight into areas such as user sentiment analysis, trend detection, for example in the case of Twitter.
\\
\\
\noindent \textbf{Relational Data}:
Relationships tend to form between i.e individuals or organisations, for example, for reasons of common interest. 
Popular examples of social ties and social interest can be seen in networks such as Facebook, Instagram, Twitter and Flickr. Analysis of network data using the vertex and edge features, often carried out for descriptive or inferential purposes, has resulted in the study of relational network science as a growing interdisciplinary field and it has found applications in a wide range of areas such as sociology \cite{wasserman},  \cite{palla}, physics \cite{adamic}, biology \cite{mich}, computer science\cite{falt}.
\\
\\
\textbf{SocialMedia Data}:
This type of data encompasses a variety of data streams connected through a network structure. It possesses many unique characteristics, given its size and the evolving dynamic content i.e video, images and text resulting in a heterogeneous mixture. For social network data, the key characteristics that are important are the connections, preferences, status, comments and tags. As the images, text, video and audio co-occur and each item (i.e image, post, video clip, audio clip) is defined by number of characteristics, learning features from such data is a non-trivial task. Further, interaction between the users through links or feedback adds another dimension to feature learning. Learning joint-embeddings of multiple data streams has resulted in a new research avenue. 
\section{GRAPH THEORY}
\subsection{Concepts}
Graph theory terminology is described to provide background for forthcoming discussions involving graph data.
A graph $\mathcal{G}$ is an ordered pair $(\mathcal{V,E})$. Set $\mathcal V$ is the \emph{vertex set} with $n \equiv |\mathcal{V}|$ denoting the \emph{order} of the graph. Set $\mathcal{E(G)}$ is the corresponding \emph{edge set}, with $e_{ij}$ as the edge between vertex $i$ and $j$.  We use the notation $\mathcal{V(G)}$ and $\mathcal{E(G)}$ to denote a vertex and an edge set of given graph $\mathcal G$.
\\
\\
\textbf{Types of Graphs}: \emph{Simple} graphs are considered throughout this paper. Simple graphs have pair of vertices connected by one edge only. Other graphs discussed in this survey are undirected, directed and weighted. \emph{Undirected} graphs have each edge as an unordered pair $\{v, w\}$. In a \emph{directed} graph, the edges are ordered pairs. In a \emph{weighted} graph, a weight function $\omega : f \rightarrow \mathcal{R}$ assigns a weight on each edge. A graph is \emph{connected} if there exist paths between all pairs of vertices. If all the vertices of a graph have the \emph{same degree} then we have a \emph{regular} graph. A \emph{complete} graph is one in which there is an edge between every pair. 
\\
\\
\textbf{Degree, Walk, Cycle, Path, Distance, Height and Depth}:
The \emph{degree} of a vertex $u$, denoted $deg(u)$, is the number of edges incident on $u$.  A \emph{walk} is a sequence of adjacent vertices and the corresponding edges, with the \emph{length} of the walk given by the number of edges included. We will sometimes denote the vertices in a walk of length $k$ as the sequence $v_0, \ldots v_k$. If $v_{0} = v_{k}$ (i.e the start vertex is equal to the end vertex) then the walk is a \emph{cycle}. A walk is a sequence of alternating vertices and edges. The term \emph{path} denotes walks where no vertex appears more than once. Moreover, the distance between two vertices, denoted $dist(u, v)$, is defined as the length of the shortest path between them. The \emph{height} of a vertex is the number of edges on the longest path top-down between the respective vertex and a leaf vertex. The \emph{depth} of a vertex is the number of edges from the vertex to the tree's root vertex.
\\
\\
\textbf{Subgraph}:
A subgraph $\mathcal{G}_{1}$ of a graph $\mathcal{G}$ is a graph $\mathcal{G}_{1}$ whose vertex and edge sets are subset of those of $\mathcal{G}$.
A \emph{clique} is a complete subgraph of a graph. A \emph{cycle} is also a connected subgraph where each vertex has exactly two neighbours and a graph that contains no cycles is a \emph{forest}. A connected forest is a \emph{tree}. A  \emph{subforest}  is an acyclic subgraph; a \emph{subtree} is a connected \emph{subforest}. 
The set of
neighbours for a given vertex $v$ is called the \emph{neighbourhood}
of $v$ and is denoted by $\mathcal{N}_{v}$. 
\\
\\
\textbf{Graph Isomorphism}:
Let $\mathcal{G}=(\mathcal{V},\mathcal{E})$ and $\mathcal{G}'=(\mathcal{V}',\mathcal{E}')$ be two graphs. $\mathcal{G}$ and $\mathcal{G}'$ are said to be \emph{isomorphic} if there exists a bijective function $f: \mathcal{V} \rightarrow \mathcal{V}'$ such that, for any $u, v \in \mathcal V$, we have that $f(u)$ and $f(v)$ are adjacent in $\mathcal G$ iff they are adjacent in $\mathcal G'$. Solving \emph{graph isomorphism} problems is relevant to machine learning as this provides a way of detecting similarities among data points. However, graph isomorphism is a challenging problem. There is no known polynomial time algorithm for graph isomorphism.  Early approaches to solve the graph matching problem proposed the use of graph edit-distances \cite{survey} as well as topological descriptors \cite{prado}. Using graph edit-distances involves counting the number of key operations that would transform graph $\mathcal{G}_{1}$ into $\mathcal{G}_{2}$ offering flexibility to assign costs; however, this approach suffers from the need to choose some optimal cost function for different operations as well intermediate steps of subgraph isomorphism. The use of topological descriptors mapping each graph to a feature vector again suffers from the loss of topological information through the transformation step.  A practical tractable alternative consists of using substructures formed from graphs that are computable in polynomial time is popularly known as the \emph{bag-of-structures} approach and it is discussed in detail in section four.

\subsection{MATRIX REPRESENTATION OF GRAPHS}
\subsubsection{Types of Matrices}
We need to work with the input representations of the matrices to generate the features. These are as follows: the \emph{Adjacency Matrix}, the \emph{Degree Matrix} and the \emph{Laplacian Matrix}. The adjacency matrix of a graph is denoted as $\mathcal{A}$ and encapsulates the whole topology of the graph in the $n \times  n$ form as below.
\begin{equation}
 \mathcal{A}_{ij}=\begin{cases}
     1, & if (i,j) \in \mathcal{A}\\
     0, & (i=j)
   \end{cases}
\end{equation}
The  degree matrix $D_{ii}$ is a \emph{diagonal matrix} where $d_{i}$ is the degree of vertex $i$.
\begin{equation}
 \mathcal{D}_{ii}=\begin{cases}
     d_{i} & (i=j)\\
     0, & (i\ne j)
   \end{cases}
\end{equation}
\noindent For an unweighted graph, the  normalised Laplacian matrix  $\mathcal L$ is of the form
\begin{equation}
 \mathcal{L}_{ij}=\begin{cases}
     -1  &  if \, (i,j) \in \mathcal{E} \\
    d_{i} & \, if \, i=j \\
     0 & \textit{otherwise}
   \end{cases}
\end{equation}
The spectral decomposition of a normalised Laplacian $\mathcal{L}$ is as follows. $\mathcal{L}$ is a symmetric positive semi-definite matrix and can take the form $\mathcal{L} = \Phi \Lambda \Phi^{T}$, where $\lambda = diag(\lambda_{1},\lambda_{2}, \lambda_{3} \ldots \lambda_{|V|})$ is the diagonal matrix with ordered eigenvalues of $\mathcal L$ as elements and $\Phi = (\phi_{1}, \phi_{2}, \ldots \phi_{|V|})$ is the matrix with ordered column eigenvectors. The \emph{spectrum of a graph} is the study of the eigenvalues of the adjacency matrix. 
\subsubsection{Relationship between the Matrices}
The  normalised form of the adjacency matrix  is $A_{\mathcal{N}} = \mathcal{D}^{-\frac{1}{2}}\mathcal{A}\mathcal{D}^{-\frac{1}{2}}$. The graph Laplacian can also be computed using the degree matrix and the adjacency matrix as $\mathcal{L} = \mathcal{D} - \mathcal{A}$. The normalised Laplacian is written as $\mathcal{L} = \mathcal{D}^{-\frac{1}{2}} \mathcal{L}\, \mathcal{D} ^{-\frac{1}{2}}$ and this follows on to
$\mathcal{L} = \mathcal{D}^{-\frac{1}{2}} (\mathcal{D} - \mathcal{A}) \mathcal{D} ^{-\frac{1}{2}}$.
%
%
%
\section{USE CASES}
Popular use cases of graph data: graph comparison, graph classification, graph clustering, link prediction, graph compression and graph visualisation are discussed.
\subsection{GRAPH COMPARISON} The task of graph comparison is to determine the dissimilarity or similarity between two graphs  through a mapping $s:\mathcal{G} \times \mathcal{G} \rightarrow \Re$. Traditional graph comparison algorithms have been classified into  set based, subgraph based and kernel based \cite{ninographlet}. In set based approach, the graph is represented using either a set of edges or a set of vertices whereas in subgraph based approach the subgraphs are extracted from graph and comparison is done using the number of matching subgraphs. Kernel method based approaches are discussed in section four.
\subsection{GRAPH CLASSIFICATION}
The graph classification problem is of two types: one entails a vertex classification problem and other entails a whole graph classification. In the whole graph classification, given a dataset $\mathcal{D}$ consisting of graphs, where each graph $\mathcal{G}_{i}$ has a vertex-edge set such as $\mathcal{G}_{i}$ = $(\mathcal{V}_{i},\mathcal{E}_{i})$, the goal of graph classification is to learn the model $f : \mathcal{D} \rightarrow \mathcal{Y}$ and classify the graphs into one or more classes.  
Each graph has a corresponding class label typically given as $\mathcal{Y}$ = $\{1 \ldots y\}$.  Data is divided into the \emph{training set} and the \emph{test set} and the trained model is evaluated on the test set. The accuracy is tested by comparing the predicted output $\hat{y}_{i}$ with the true class label $y_{i}$. 
\subsection{LINK PREDICTION}
A priori, we do not know which links are missing or which future links are going to be formed. The high-level view of the task of link prediction is to predict how the structure of the network evolves as existing participants form new links, break links \cite{linkmob}. For example, in protein-protein interaction networks, link prediction can identify novel interactions between proteins \cite{pplink}. 
Following the definition in \cite{linkpredsurvey}, given a network graph $\mathcal{S}$ = $\mathcal{(V,E)}$,  the task of link prediction is defined as follows. Consider $\mathcal{U}$ to be a universal set containing $\frac{|\mathcal{V}|(|\mathcal{V}|-1)}{2}$ possible links where $|\mathcal{V}|$ denotes the number of elements in the set.
Therefore, the task of link prediction is to find the links in the set $\mathcal{U}-\mathcal{E}$.  The dataset is divided randomly into two parts $\mathcal{E}^{T}$- training set and the $\mathcal{E}^{P}$- test set.
Globally, the \emph{Network Growth Prediction} problem is described as an extension of the link prediction problem. In social network analysis, it is used to predict user preferences for forming new friendships resulting in the user's social network growth \cite{linkpred}.
\subsection{GRAPH CLUSTERING}
In graph clustering, the edge structure plays an important role. The vertices  of the graph are grouped into clusters in such a way that there are many edges \emph{within} the formed cluster and comparatively \emph{fewer} between the clusters \cite{graphclustering}. There are two main approaches: \emph{within-graph} clustering and \emph{between-graph} clustering. As the name implies, within-graph clustering methods divide the vertices within a graph into clusters, whereas in the case of between-graphs, the clustering algorithm works on a set of graphs which are divided into different clusters. In biology, the applications of graph clustering are in gene regulatory networks, metabolic networks, neural networks and food webs. In social networks, clustering is an approach used for community detection, for example in \cite{socialcircle} the authors attempt to identify a user's circles, each a subset of her friends and this problem of circle detection is formulated as a clustering problem. Clustering is also used to identify communication networks, organisational networks, online communities and this is finely discussed in \cite{cdg}.
\subsection{FURTHER USE CASES}
Large scale graphs, such as web or social media graphs typically contain more more than a billion edges and are growing quickly. Learning from large graphs is extremely challenging from a computational viewpoint. Two use cases have gained recent traction to address this challenge: Graph Compression and Graph Visualisation. 
A compressed representation of a graph encodes its topological structure \cite{ferder}. Constructing a good graph representation is a space-saving approach and several compression schemes have been researched to propose various graph representations \cite{peternek}. Graph visualisation explicitly shows us the connections between vertices, communities or subgraphs. 
The visual graphic of a graph can show some interesting properties to enable the reader to study the network from another angle. Some interesting visualisations can be seen in \cite{deepwalk}, \cite{sdne} and \cite{line}. Nevertheless, challenges of customisability, layout and generating 
dynamic visualisations remain an ongoing case to solve.  
\section{KERNEL METHODS}
Kernel methods  are a widely used class of algorithms that could be applied to any data structure. Kernel approaches are also used as building blocks in some representation learning methods described in the following section. A \emph{kernel function} is the inner product of two vectors in feature space. It \emph{isolates} the \emph{learning algorithm} from the instances. This means that the learning algorithm relies exclusively on the kernel values between the instances without the need to explicitly use the original data representation.
\\
\\
Formally, let $\mathcal{X}$ be a non-empty set and let $k: \mathcal{X} \times \mathcal{X} \rightarrow \mathcal{R}$, where $\times$ denotes set product. Kernel $k$ is symmetric if $k(x,y)=k(y,x)$ and $k$ is positive definite given that $n \ge 1$ and $x_{1} \ldots x_{n} \in \mathcal{X}$ and matrix $k$ is defined by 
$k_{ij} = k(x_{i},x_{j})$ is positive definite i.e we have 
$\sum_{i,j=1}^{n}c_{i}c_{j}k(x_{i},x_{j}) \ge \,0 \, \, \forall$ $c_{1} \ldots c_{n} \in \Re$. An alternative way of writing a kernel function is $k(x,x') = <\phi(x), \phi(x')>$, where $\phi(x)$ is a feature vector. In the scope of this survey, kernel methods for graph-structured data are discussed.

 
\subsection{KERNELS METHODS FOR GRAPHS}
Learning dictionaries of structured data is an approach that started in the late 1990s. The \emph{Bag-of-Structures} approach is one in which each data point is a derived vector representation for a given type of graph substructure. Using the bag-of-structures approach, the feature representation for each kernel type is fixed and each dimension corresponds to a type of substructure. This results in very high dimensional kernel space. 
Formally, let $\mathcal{G}$ be a non-empty set of graphs, then kernel $k: \mathcal{G} \times \mathcal{G} \rightarrow \mathcal{R}$ is called a \emph{graph kernel}, here $<\phi (\mathcal{G}), \phi (\mathcal{G}')>$ are the respective feature vectors.  
\begin{equation}
k(\mathcal{G},\mathcal{G}')= <\phi (\mathcal{G}), \phi (\mathcal{G}') > 
\end{equation}
Existing graphs kernels are an instance of $\mathcal{R}$-convolution kernels. The $\mathcal{R}$-convolution framework is formed on pairs of graphs following the decomposition of two structured objects.
\cite{haussler} proposed the idea of decomposing an object into atomic substructures. Each new decomposition relation $\mathcal{R}$ results in a new graph kernel. 

\begin{equation}
k_{conv}(x,x^{'})= \sum_{(x_{d},x )\in \Re}\sum_{(x^{'}_{d},x^{'}) \in \Re}\mathcal{K}_{parts}(x_{d},x^{'}_{d})
\end{equation}
There are two fundamental learning approaches when working with graph kernels: learning \emph{Kernels defined on Graphs} and learning \emph{Kernels defined between Graphs}. Kondor and Lafferty \cite{lafferty} proposed the idea of Kernels on Graphs which are kernels formed between vertices of a single graph. This was further extended by Smola and Kondor \cite{smola}. The Kernel between graphs approach was proposed by Gartner \cite{gartner}. 
We review some of the kernels using the bag-of-structures approach and these are categorised into three major families: \emph{Kernels on Walks and Paths}, \emph{Kernels on Subtrees} and \emph{Kernels on Subgraphs}. However, there are other methods used to derive graph features that do not depend on the bag-of-structures approach and these are discussed towards the end of this section.  

\subsubsection{Kernels on Walks and Paths}
\textbf{Random Walk Kernels} proposed by Gartner, are based on counting the number of walk based substructures formed from the sequences of vertices between graphs in dataset $\mathcal{D}$. To find common walks in two graphs, a product graph is used. A product graph is formed from identically labeled vertices and edges from $\mathcal{G}_{1}$ and $\mathcal{G}_{2}$. 
Here ($p_{1}$,$p_{2}$) are the starting probabilities and ($q_{1},q_{2}$) the stopping probabilities of the random walk.
$\mathcal{A}_{1},\mathcal{A}_{2}$ are the adjacency matrices of $\mathcal{G}_{1}$ and $\mathcal{G}_{2}$. The number of length $l$ common walks on the direct product graph $\mathcal{G}$ where $\otimes$ is the Kronecker product of two matrices is computed as follows. 
\begin{equation}
(q_{1} \otimes q_{2})(\mathcal{A}_{1}^{T} \otimes \mathcal{A}_{2}^{T} )(p_{1} \otimes p_{2})
\label{eq:rw_kernel}
\end{equation}
Formally, the random walk kernel between two graphs can be defined as follows. 
\begin{equation}
\begin{split}
k(\mathcal{G}_{1},\mathcal{G}_{2})=\sum^{\infty}_{l=0}\,(q_{1}\otimes q_{2})
(\mathcal{A}_{1}^{T} \otimes \mathcal{A}_{2}^{T})^{l} (p_{1}\otimes p_{2}) (q_{1}\otimes q_{2})\\
(I - \lambda(\mathcal{A}_{1}^{T} \otimes \mathcal{A}_{2}^{T}))^{-1}(p_{1} \otimes p_{2})
\end{split}
\end{equation}
Here, $\lambda$ is the discounting factor applied to long walks, with all the common walks summed for all different lengths.
In a neater form, the random walk kernel is defined as follows where $q=(q_{1}\otimes q_{2})$ and $p=(p_{1}\otimes p_{2})$.
\begin{equation}
\begin{split}
k({G}_{1},\mathcal{G}_{2})=q^{T}(1-\lambda \mathcal{A})^{-1}p
\label{eq:rwneat}
\end{split}
\end{equation}
For the random walk approach, the runtime is $\mathcal{O}({n^6})$.  Artificial kernel values that are a consequence of repeated vertex-edge visits. This is called \emph{tottering}, a phenomenon in which different walks include repetitions of vertex and edges resulting  in a high structural similarity score as the same vertex or edge is repeatedly counted. Another phenomenon, \emph{halting} is associated with random walk kernels where a weight factor $\lambda$ down weights the longer walks. 
Random walk kernels have been extensively studied  with suggestions for improvement such as fast computation of random walks, label enrichment and methods to prevent tottering \cite{graphfeaturemaps}, \cite{fastrandomwalk}.
\\
\\
\textbf{Shortest Path Kernel} \cite{shortpath} is computed by computing all pairs of the shortest-path $p$, for given length $n$, for each graph in dataset $\mathcal{D}$. Given the shortest paths,  $p$  and $p'$ for the graphs $\mathcal{G}$ and  $\mathcal{G}'$, the kernel is formed as the sum over all pairs of edges $\mathcal{E}_{p}$ and $\mathcal{E}_{p'}$ from $p$  and $p'$  using a sensible choice of kernel on the edges. 
\begin{equation}
\mathcal{K}(\mathcal{G},\mathcal{G}') =  \sum_{p \in \mathcal{E} p}\sum_{p' \in \mathcal{E}_{p}} \mathcal{K}^{n} (\phi(\mathcal{G}_{p}),\phi(\mathcal{G}'_{p}))
\end{equation}
To overcome the problems associated with walk kernel methods that suffer from tottering and halting, the approach is to define kernels on paths. However, computing all paths is intractable, while computing all pairs-shortest-paths is $\mathcal{O}(n^{3})$ and comparing all pairs shortest paths is $\mathcal{O}(n^{4})$. Considering large graphs, this becomes computationally expensive.
\\
\\ 
\textbf{Cyclic Pattern Kernel} \cite{cyclic} is  computed by counting the number of common cycles that appear in each of the graphs in $\mathcal{D}$. Formally, it is defined as follows where $\phi(\mathcal{G})$ of the feature of the graph.
\begin{equation}
\mathcal{K}(\mathcal{G},\mathcal{G}') = \mathcal{K}(\phi(\mathcal{G}_{cp}),\phi(\mathcal{G}'_{cp}))
\end{equation}
However, to compute the cyclic pattern kernels, the cyclic and tree patterns from each graph need to be computed and then a form of intersection is applied. Finding all the cycles in a graph in polynomial time is a challenge and since there is no known algorithm to compute this, researchers have resorted to sampling and time-bounded measures.
\subsubsection{Kernels on Subtrees}
\textbf{Subtree Kernel}, proposed by Ramon and Gartner \cite{expressivity}, is computed by finding the common subtrees in each of the graphs in dataset $\mathcal{D}$ and comparing them. By definition, a \emph{subtree} of graph $\mathcal{G}$ is a connected subset
of distinct vertices of $\mathcal{G}$ with an underlying tree structure. Finding common tree-like neighbourhoods amongst graphs in $\mathcal{D}$ amounts to counting pairs of identical subtrees with given height $h$. The advantage is that we have a richer representation of the graph structure encapsulating its topology. Subtree kernels on graphs is the sum of subtree kernel on vertices.
\begin{equation}
\mathcal{K}(\mathcal{G},\mathcal{G}') = \sum_{a \in \mathcal{V(G)}}\sum_{b \in \mathcal{V(\mathcal G')}}\mathcal{K}_{h}(a,b)
\end{equation}
However, the runtime is affected by the recursion depth of the subtree like patterns and therefore subtree kernels also suffer from \emph{tottering}. Computing a subtree pattern will have a signature represented by the sequence of labels of the vertices in the sequence \cite{fastsubtree}. 
\\
\\
\textbf{Weisfelier-Lehman Kernel} 
\cite{lehman} (WL) is a fast subtree kernel. It uses the Weisfelier-Lehman isomorphism test composed of the steps of iterative multi-set label determination, label compression and relabelling. Here, $h$ is the depth, with $l$ given to be a relabelling function, the WL kernel is defined as follows
\begin{equation}
k(\mathcal{G},\mathcal{G}') = \sum^{h}_{i=0} k (l^{i}(\mathcal{G}),l^{i}(\mathcal{G}'))
\label{eq:14}
 \end{equation}
The 1-d WL algorithm is a type of colouring scheme. It starts with iterating through each vertex label and its neighbouring set. Each vertex is relabelled with the compressed label that is used at the next iteration. 
The algorithm goes through $k$ iterations using the compressed labels to construct a frequency vector for each graph. The recolouring of the vertices converges when the number of distinct colours stop increasing, which means that the vertices in the graph cannot be further partitioned.  The output  
contains the frequency of compressed labels occurring in $k$ iterations. However, the WL kernel does exact matching and therefore the labels \emph{EFGH} and \emph{EFGJ} return a zero similarity match and so it fails to take partial similarity into account. Nevertheless, it scales well to large, labeled graphs.
\subsubsection{Kernels on Subgraphs}
\textbf{Subgraph kernel} is computed using the idea  that similar graphs tend to have similar subgraphs which could be used for graph comparison. Connected non-isomorphic subgraphs of size $k$  are referred to as \emph{graphlets}. A $k$-sized graphlet is defined as $\mathcal{G}_{k} = \{ g_{1}, g_{2}, g_{3} \ldots g_{n_{k}}\}$ where $n_{k}$ is the unique number of graphlets of size $k$. Following  \cite{ninographlet}, let $\phi(\mathcal{G}_f)$ be the normalised vector of length $n_{k}$ whose $i^{th}$ element is the frequency of the graphlet $g_{i}$ in $\mathcal{G}$ and let $s_{j}$ denote the number of times $g_{k    }$ occurs as a subgraph of $\mathcal{G}$. That is, 
\begin{equation}
\phi(\mathcal{G}_f) = \Big (\frac{s_{1}}{\sum^{n_{k}}_{j}S_{1}} \ldots \ldots \frac{s_{j}}{\sum^{n_{k}}_{j}S_{j}} \Big )
\end{equation}
Graphlet kernels compute the similarity between two graphs using the dot product of the count vectors of all possible connected subgraphs of some order $k$. 
\begin{equation}
\mathcal{K}(\mathcal{G},\mathcal{G}') = \big (\phi(\mathcal{G}_{f}),\phi(\mathcal{G}'_{f})) 
\end{equation}
\emph{Weighted Decomposition Kernel} \cite{menchetti}, applied to the use case of protein sequence and molecule graph classification  uses a subgraph $s$ from graph $\mathcal{G}$ called \emph{selector}, with associated kernel $\delta$, weighted according to the  match within a \emph{context of occurrence} i.e a set of subgraphs, $z = (z_{1}.....z_{D})$ with associated kernel $k$. Substructures are matched according to an equality predicate measure and the kernel is computed as follows.
\begin{equation}
\mathcal{K}(\mathcal{G},\mathcal{G}') = \sum_{(s,z) \in R^{-1}(G),(s^{'},z^{'}) \in R^{-1}(G^{'})} \delta(s,s^{'})\sum_{d=1}^{D}k(z_{d},z^{'}_{d})
\end{equation}
\subsubsection{Challenges: Working with Bag-of-Structures approach}
\textbf{Diagonal Dominance}: The bag-of-structures approach recursively decomposes structured objects into substructures but this results in various challenges. For example, one of the challenges is the \emph{Diagonal Dominance} problem, where the kernel matrix becomes closer to the identity matrix. This happens when different substructures are regarded as different features and as these substructures grow in number, the feature set grows larger. Therefore, the probability that given two graphs will contain similar substructures diminishes. Hence, the highest similarity of a graph is to itself as compared to other graphs within the training set. 
\\
\\
\textbf{Substructure Sparsity} $\&$ \textbf{Substructure Dependence}: Other practical issues include \emph{Substructure Sparsity}, the problem of  where only few of the substructures are common across the graph. \emph{Substructure Dependence} is the problem where subgraphs that occur are not independent, as one subgraph could be found inside another or could be arrived at by making modifications to vertices and edges of other subgraphs. Therefore, features that are represented by these subgraphs turn out to be similar in nature.
Finally, most graph kernels consider each substructure as separate feature and this not only increases the feature set but also results in similar features. Therefore, the substructures that occur frequently, those that often encompass the lower order substructures, tend to dominate the occurrence index. 
\subsubsection{Other Approaches}
Intensive research on kernels for structured data with associated applications can be found in \cite{watkins},\cite{jaakkola}, \cite{leslie}, \cite{lodhi}, \cite{collins}, \cite{koyanagi}, \cite{surveygartner}. Nevertheless, there are other kernel approaches that work equally well on structured data and are used to derive graph features. For example, using hash kernels for structured data \cite{hashkerns} results in hashing that preserves information and also facilitates dimensionality reduction. Neighbourhood hash kernel \cite{hashkernel} is one such example that uses hashing techniques for labeled graphs encoding the vertex neighbourhood and topology information using the bit arrays and logical operations. Each vertex label is transformed into a bit label with a mapping function. Using the $XOR$ operation on bit labels for a given vertex label around the neighbourhood of that vertex results in a unique encoding for that vertex and its neighbourhood. Each of these unique encodings combined into a feature matrix, are used to learn the whole graph representation.
The use of heat kernel for generating graph representations has 
 has found numerous uses. For example, authors Xiao and Handcock, explore numerous ways to compute the heat kernel for graph clustering \cite{handcock}. Many kernels designed for structured data have leveraged the use of probabilistic graphical models to discriminate features. Consider 
the Fisher Kernel \cite{fisher} which compares two objects by fitting a generative model to the entire dataset and then using the fisher information matrix and the fisher score for each data point defining the kernel in this manner. Another example, the Probability Product Kernel \cite{jebbara} is based on the central idea to define kernels between distributions. Themes of such existing methods have been incorporated into recent approaches for learning graph representations.
\section{LEARNING GRAPH REPRESENTATIONS}
Five taxonomies are proposed by categorising them according to a set of baseline techniques used to construct the methods in each of the following approaches. These are: Kernel  approaches, Convolution approaches, Graph neural network approaches, Graph embedding approaches and Probabilistic approaches. The term \emph{graph representation} is defined as a \emph{learned} graph feature which is obtained following neural network computation and each \emph{learned} representation encodes the respective topological information about the graph.
\subsection{KERNEL APPROACHES}
Recent advances have highlighted the relation between  neural networks and kernel methods. For example, Cho and Saul \cite{cho,cho1} construct kernels that mimic neural networks whilst Mairal et al \cite{mairal}, show the connection between convolutional neural networks and kernels. Further references are  \cite{zhang},\cite{tamir} and \cite{teh}. Kernel approaches are characterised by the use of kernel methods for graph data incorporating neural learning. 
\\
\\ 
\textbf{Deep Graph Kernels}: \cite{deepgraphkernels} is one of the foremost approaches to combine graph kernels with deep learning techniques championed by Yanardag and Viswanathan. They tackle the challenge of capturing meaningful semantics between substructures. The bag-of-structures approach suffers from the issues of substructure dependence, substructure sparsity and diagonal dominance (section 4.1.4). The authors alleviate these by introducing the encoding matrix $\mathcal{M}$, a $|\mathcal{S}| \times |\mathcal{S}|$ positive semidefinite matrix that \emph{encodes} the relationships between the substructures where $|\mathcal{S}|$ is the size of vocabulary of substructures extracted from the training data. This is achieved by designing $\mathcal{M}$ such that it respects the similarity of the substructure space.  The kernel is then  defined as
\begin{equation}
\mathcal{K}(\mathcal{G},\mathcal{G}')= \phi (\mathcal{G})^{T}\mathcal{M}\phi (\mathcal{G}') 
\end{equation}
Approaches to calculating $\mathcal{M}$ are as follows, first by using the edit-distance relationship between the substructures and second by learning the latent representations of substructures using probabilistic neural language models (section 5.3.2). Data corpuses are generated such that a co-occurrence relationship is
partially preserved. The neural language model is built using the continuous bag-of-words or Skip-Gram and  trained using hierarchical softmax. Interesting performance results for deep graph kernels are shown on social network and bioinformatics dataset. 
\\
\\
\textbf{Kernel Neural Network}: In \cite{knn}, the authors leverage kernels defined over structured data such as sequences and graphs to derive neural operations. They design a new architecture using kernel inner product, embedding it into a recurrent neural network. Within the scope of this review, one such example is explained to illustrate the embedding of the graph kernel into the neural module. Given, the random walk kernel (\ref{eq:rw_kernel}) concerning feature
vectors $f_x$, the kernel and neural computations are linked as follows.
\begin{equation}
\begin{array}{rcl}
c_{1}[v] &=& w^{1}f_{v}\\
c_{j}[v] &=& \lambda \sum_{u \in N(v)} c_{j-1}[u] \odot W^{j}f_{v}, \quad 1 < j \le n
\end{array}
\label{eq:rw_neural}
\end{equation}
Here $\odot$ is the element wise product, $N_{(v)}$ represents the neighbourhood of the graph around vertex $v$, $W$ is the weight matrix and $c_{j}[t]$ and $h[t]$ are the pre and post activation states.
Equation (\ref{eq:rw_neural})  provides a model that embeds the random walk kernel into the neural framework. Here, $c_{\ast}[v]$ is the random walk count vector for a given vertex $v$ and ${h}_{\mathcal{G}}$ is the latent representation of the graph aggregated from vertex vectors and this learned representation could be used for either classification or regression.
\begin{equation}
h_{\mathcal{G}} = \sigma(\sum_{v}c_{n}[v]) \quad 1 <j \le n
\end{equation}
In this manner, illustrating with the random walk kernel example, the neural module embeds sequence similarity within the architecture. Kernels could be used for the single and multiple layer constructions in which case through stacking the output states $h^{l}[t]$ of the $l^{th}$ layer are fed into the $(l+1)$ layer as the input sequence. The authors derive similar templates for a variety of graph kernels families supporting their theory with numerous experiments. 
\subsection{CONVOLUTION APPROACHES} Early researchers 
Fukushima \cite{neo}, Atlas \cite{atlas}, LeCun \cite{lecunn}  contributed to the development of convolutional neural networks (CNNs). Recent contributions by Mallet \cite{mallet} and Wu  \cite{wujx} focus on the CNN theoretical framework.
CNN architectures are able to extract  representations from data that have an underlying spatiotemporal grid structure, making them suited for working with image, 
video \cite{spatio} and speech data \cite{acoustic}. CNNs are designed to extract local features across the signal domain by extracting the local stationarity property of the input data. 
\\
\\
\textbf{Learning with CNNs} involves two key operations \emph{convolution} and \emph{pooling}. Localised convolutional filters learned from data identify similar features. Convolutional filters are shift invariant and location independent hence recognising identical features independently. For example, with image data the CNN convolution operator takes an input and convolves kernel filter over it using stride $s$ and appropriate parameters for tuning. Feature maps generated as a result of the convolution operator are then fed into pooling layer to provide a compressed output. Further a fundamental concept for designing convolutional neural network architectures uses a \emph{receptive field} which in essence is the \emph{local region} of the input. Selecting vertices of a graph for creating a convolution is similar to selecting receptive field in a classic neural network. Addressing the practical concern in the increase of the number of parameters when features in one layer are connected to the features in the layer beneath, Coates et al \cite{receptive} propose connecting each feature extractor to a \emph{local region} i.e the receptive field of inputs. Features are grouped together based on a similarity measure to limit the number of connections between the two layers.
\\
\\
\textbf{Convolutions on graph}: Many data have irregular structure in their underlying graph, due to the underlying irregular spatial geometry, such data is known as \emph{non-euclidean} data. Regular, lattice type, underlying structure is found in  time-series and image
data whereas irregular underlying structure is found in text data, sensor data, mesh data, social network data and gene data, for example. To design convolutional networks for graph data, we need to use a similar convolutional operator but \emph{one that works on graph data domain, one that works on an irregular domain}. 
\\
\\
We define concepts that are used to formulate the graph convolution operator in the papers discussed below. A \emph{graph signal}, considering undirected graphs, is a function mapping $x: \mathcal{V} \rightarrow \Re $ defined on the vertices of the graph and represented by vector $x \in \Re^{N}$, where the $n^{th}$ component of the vector $x$ represents signal value at the $n^{th}$ vertex in $\mathcal{V}$. One can think of the data as tied to the vertex of the graph, for example a vertex could denote a single gene in a gene-gene interaction network.
\\
\\
Classical fourier transform of a function $f$, frequency $w$ is the inner product of $f$ with eigenfunction $exp(2\pi i wt)$.
\begin{equation}
\widehat{f(w)} = \int_{-\infty}^{\infty}f(t)\,exp(-2\pi i wt)\, dx
\end{equation}
The graph fourier transform of a function $f: \mathcal{V} \rightarrow \mathcal{R}$ is the expansion of $f$ in terms of the eigenfunctions of the graph Laplacian.
$\mathcal{L}$ is positive semidefinite, has $\{u_{l}\}^{n-1}_{l=0} \in \Re^{n}$ orthonormal set of eigenvectors and the nonnegative eigenvalues $\{\lambda_{l}\}^{n-1}_{l=0}\in \Re^{n}$ 
 and the eigendecomposition as $\mathcal{L} = \mathcal{U}\,\mathcal{\lambda}\,\mathcal{U}^{T}$ where $\lambda = diag[\lambda_{0}.....\lambda_{n-1}]$, $\mathcal{U}$ the fourier basis. 
%
%
Fourier transform converts a signal from a time domain into the frequency domain. 
The graph fourier transform $\hat{x}$ of a spatial signal $x$ is $\hat{x}=\mathcal{U}^{T}x \in \, \Re$ followed by the inverse $x=\mathcal{U}\hat{x}$ \cite{emerge}.
%
Formally, \emph{convolution} is defined through an integral that expresses the amount of overlap of given function $g$ as it is shifted over another function $f$. Mathematically, it is written as follows \cite{bracewell}.
\begin{equation}
f \ast g = \int^{+\infty}_{-\infty}f(\tau)g(t-\tau)d\tau = \int^{+\infty}_{-\infty}g(\tau)f(t-\tau)d\tau
\end{equation}
Convolution is a linear operator that diagonalises the fourier basis and as we can express a meaningful translation in the fourier domain instead of the vertex domain. The convolution operator on the graph is defined as follows, where $\odot$ is the element wise product.
\begin{equation}
x_{\mathcal{G}} \ast y = \mathcal{U}((\mathcal{U}^{T}x)\odot(\mathcal{U}^{T}y))
\end{equation}
Discretised convolutions used in CNN, commonly for image data are defined on regular grids both 2D and 3D and hence not applicable to the graph data domain. For irregular grids such as graphs, we need to define localised filters and these are known as \emph{spectral graph convolutions}. Our interest is to obtain spectral convolutions on graphs. Spectral graph convolutions exploit the fact that convolutions are multiplications in the fourier domain.
A spectral convolution of the signal $x$ with a filter $g_{\theta}$ as follows. Here, $g_{\theta}(\lambda)= diag(\theta)$, where $\theta \in \Re$ is vector of fourier coefficients. 
\begin{equation}
\begin{split}
y = g_{\theta} \, (\mathcal{L}) \, x 
  = g_{\theta} \, (\mathcal{U} \,\mathcal{\lambda}\, \mathcal{U}^{T}) \,x 
  = \mathcal{U} \, g_{\theta}(\lambda) \,\mathcal{U}^{T} \, x
\end{split}
\label{eq:24}
\end{equation}

\subsubsection{Spatial and Spectral Approaches}
Two main methods based on convolutional approaches as proposed in literature for graph data. These are defined as the \emph{spatial approach} and the \emph{spectral approach}. Spatial approach is characterised using the notion of localised receptive fields formed by the neighbourhood of a vertex in the context of graph data for CNNs. The receptive fields are formed as a direct measure of distance in a graph, where given a vertex considered to be the center of the filter, we look around at vertices within a particular number of hops away. Spectral approach is characterised by using measures of distance based on decompositions of the graph Laplacian. Both approaches require careful consideration, creating spectral convolutions is dependent on graph structure. For spatial convolution, there is the need to create shift-invariant convolutions for graph data, as well as the problem specific need to determine vertex ranking and neighbourhood ordering. Another drawback observed with CNNs is that convolution operations are only applied to vertex features assuming that the graph domain is fixed but in many cases the graph can be noisy and some graphs are computed \emph{a priori}, this is not necessarily reflects the relationships between the actors. 
\subsubsection{Spatial Approach}
\textbf{Spatial Convolutions}: The use of spatial approach can be noted in {\small{PATCHY-SAN}}(PS)\cite{learningCnn}. Here, it is used to learn the graph representations using a CNN in a supervised fashion creating receptive fields in the manner similar to how classical CNNs work on images. PS uses a number of steps to create graph derived receptive fields. During vertex sequence selection of a section, for a given graph $\mathcal{G}$, a sequence of vertices is identified and in the neighbourhood assembly step the neighbours are identified for creating receptive fields. Thus the receptive field for a given vertex results in a \emph{neighbourhood} receptive field. Following the creation of a neighbourhood receptive field, a normalisation procedure is implemented which in essence is a form of vertex ordering to create a vector in vector space for deriving graph features used to learn graph representations.
\cite{dcnn} is another example of spatial approach, using random walk (\ref{eq:rw_kernel}) to select spatially close vertices. It forms the convolutions by associating the $i^{th}$ parameter $w_{i}$ with the $i^{th}$ power of the transition matrix $(P^{i})$. Using the transition matrix results in $\mathcal{O}({N}^{2})$ complexity and \cite{sparsedcnn} suggests a simple method for thresholding graphs resulting in optimum memory reducing complexity to $\mathcal{O}({N})$.
\\
\\
\textbf{Generalised Convolutions}: \cite{gencnn} take a generalised approach to CNNs, allowing for generating convolutions on graphs of different sizes. This proposition uses a spatial approach based on a random walk based transition matrix on the graph to select ranked $k$ neighbours. The transition matrix $P$ is then used to derive $\mathcal{Q}^{k}_{ij}$ which calculates the expected number of visits from a given vertex $X_{i}$ to given vertex $X_{j}$ in $k$ steps. The convolution over the graph vertex $X_{i}$ is represented as a tensor product of the form $(M,N,d) \rightarrow (M,N,p,d)$ where the $4D$ tensor includes the top $p$ neighbours of each feature selected by $Q^{k}$, for $M$ observations, $N$ features at depth $d$. The authors test the approach on the Merck molecular activity Kaggle challenge.
\\
\\
\textbf{Motif CNN}: Motif are small subgraphs patterns demonstrating specific connections amongst vertices. \cite{motifcnn} present a motif-based CNN where the motifs are defined to create a receptive field around a target vertex of interest. A motif-convolutional unit at vertex $v_{i}$ is defined such that the features of all vertices that are locally connected through that motif are weighted according to their respective semantic roles. \emph{motif-cnn} uses an attention mechanism to integrate features extracted from multiple motifs for semi-supervised learning. To the tackle the explicit assumption used by the laplacian for undirected graphs i.e symmetric laplacian matrix, MotifNet uses motif induced adjacencies by constructing a symmetric motif adjacency matrix. \cite{motifnet}.
\\
\\
In \cite{geometric}, Bronstein et al, discuss non-euclidean data, coined under the umbrella term \emph{Geometric deep learning} detailing the learning problems. \cite{geometricmatrix} treat matrix completion problem, using spatial patterns extracted by the CNN architecture designed to work on multiple graphs. The column graph is thought of as a social network capturing relations between users and similarity of their tastes whereas row graph represents the similarities of the items items. 
\subsubsection{Spectral Approach}
\textbf{Spectral Networks} are introduced by \cite{locally} and show a construction technique for connecting locally formed neighbourhood graphs. The idea behind using spectral networks is to generalise the convolutional network through the graph Fourier transform. 
For spectral construction, the spectrum of the graph Laplacian is exploited to generalise the convolution operator. Each of the constructions are tested on variations of the MNIST dataset. 
In \cite{deepgsd}, authors build on the above work using spectral graph convolution filters  (\ref{eq:24}). The authors train a \emph{graph convolution layer} performing forward and backward pass given a Fourier matrix $U$, an interpolation kernel $\mathcal{K}$ and weights $w$. During the forward and backward pass the tasks amounts to learning spectral filters on the graph.
In the first variation, coordinates are extracted from subsampled MNIST data, forming convolutions via the Laplacian spectrum. In the second variation, MNIST data are projected onto 3D sphere. 
\\
\\
\textbf{Applications to molecular data}. \cite{neuralfingerprints} shows a special case of an application for predicting properties of molecules  described as graphs. Using a custom-made CNN, a local filter is applied to an atom and its neighbourhood. Molecular graphs, representing the properties of the original molecules  are of particular interest in predicting properties of molecules. State of the art method, Quantitative Structure Activity Relationship (QSAR) \cite{qsar} uses circular fingerprints with a fully connected neural network. The authors use existing methods to create circular fingerprints to construct a differentiable fingerprint changing each non-differentiable operation to a differentiable one. A differentiable fingerprint is created inputting a molecular graph resulting in a neural fingerprint. 
Differentiable fingerprints can be optimised to include only relevant features and using similarity between fragments, neural fingerprints become more meaningful. In the spirit of neural fingerprints, molecular graph convolutions \cite{mole} is another architecture proposed to learn from undirected small graphs of molecules. In \cite{adaptivegraphs}, the authors show how to use original graph data consisting of diverse graph structures by constructing a customised graph Laplacian that uses a filter by combining the neighbourhood features according to the unique graph topology. \cite{generativeconv} adapt the use of convolutions with auto-encoders to reconstruct/learn the latent/damaged fingerprint representations.
\\
\\
\textbf{Generalising CNNs to Graphs}:\cite{fastlocal} provides a method to generalise CNNs to graph data using thorough spectral theoretical formulation. 
During the feature extraction phase, the authors perform graph signal filtering and graph coarsening. In the graph filtering phase, following a spectral formulation, strictly localised filters are defined within $k$ radius ball. In the graph coarsening phase, Graclus \cite{graclus} a fast graph clustering software. It computes the normalised cut and ratio association for a given undirected graph without any eigenvector computation \cite{gra}. As pooling compresses the output, meaningfully arranged graph neighbourhoods need to be defined. To do this optimally, the authors devise an efficient binary tree arrangement of the vertices resulting in a strategy similar to constructing a 1D signal. 
\\
\\
\textbf{Graph Convolutional Network}:\cite{gcn} introduces the Graph Convolutional Network (GCN)  for vertex classification in a semi-supervised setting. A neural network model $f(X,A)$ is used to encode the graph structure using a layer-wise propagation rule where features $X$ were derived using the popular Weisfeiler-Lehman algorithm (\ref{eq:14}) for adjacency matrix $A$. 
Given that evaluating the spectral convolution $g_{\theta} \ast x = Ug_{\theta}U^{T}x$, where $x$ is a signal for every vertex, would be computationally expensive, particularly for large graphs, following Hammond et al \cite{hammond}, $g_{\theta}$ could be approximated by Chebyshev polynomials $g_{\theta'}(\lambda) \approx \sum^{K}_{k=0}\theta'_{k}T_{k}(\widetilde{\lambda})$. 
\cite{populationgraph} shows an application of GCN in population graphs for brain analysis combining imaging and non-imaging data.
Modelling relational data stored in knowledge bases is useful in a number of tasks such as question answering, knowledge completion and information retrieval.\cite{rcgn} propose a relational graph convolution network (R-GCN), an extension of GCN, for the task of link prediction, predicting facts and entity classification. The network consists of an encoder a R-GCN that produces latent representations of the entities and a decoder which is a tensor factorisation model. 
\\
\\
\textbf{Recognition and Attention Mechanism} (GAT). In \cite{beyondgrids}, 2D feature maps are transformed into a graph structure where vertices define regions and edges capture the relationship between regions. Using the steps of \emph{graph projection}, \emph{graph convolution} and \emph{graph re-projection}, context modelling and recognition is done with graph structure. \cite{graphattention} use attention based architecture to perform vertex classification of graph structured data. It computes the \emph{importance} of each edge by processing only features of its incident vertices. This model is then applicable for inductive learning where it can be generalized to unseen graphs.
\subsection{GRAPH NEURAL NETWORKS}
We define a Graph based Neural Network (GNN) framework as one in which the connectivity among units follow the graph $\mathcal{G}(\mathcal{V},\mathcal{E})$ structure.
\\
\\
\textbf{Graph Neural Network} (GNN):\cite{graphnn} is one of the earliest approaches to propose a neural network architecture motivated by graph structure. A state vector $x_v$ is attached to each vertex $v$ given the information contained in its neighbourhood, where each vertex contains vertex-level label information $l_v$. Two main steps form the GNN. First, a parametric \emph{transition function} $x_v=f_{w}(l_v, l_{N(v)})$ (which expresses the dependence between a vertex $v$, its label $l_{v}$ and its neighbourhood $N(v)$) propagates the learned vertex representations. Second, the local output function $O_v = g_{w}(x_v, l_v)$ maps vertex representations to respective graph labels. The encoding network is built by storing the states of each vertex and updating the state when activated. GNN learns in a recursive manner, with a recurrent relation used to obtain the embedding $x_v$ for each vertex $v$ in the euclidean space. The model is differentiable end-to-end with learning parameter $w$ tasked as a minimisation of a quadratic cost function.
\\
\\
\textbf{Graph Gated Sequence Neural Network and Gated Graph Transformer Network}: \cite{ggnn} is a modification of GNN functionality. The graph gated sequence neural network (GGSNN) allows for non-sequential outputs. A classic example of this is outputting logical formulas. Using Gated Recurrent Units (GRUs)\cite{gru}, GGSNN unrolls a recurrence for a fixed number of steps. Backpropagation through time is used to compute the gradients. The propagation model 
is in the same spirit of GNN, with each vertex representation updated following a recurrence relation. 
The output model is defined per vertex and is a differentiable function 
mapping to an output. 
\cite{graphstate} proposes the use of Gated Graph Transformer Neural Network (GGT-NN) in order to solve reasoning problems. A number of graph transformations such as node addition, node state update, edge update, propagation and aggregation are combined to solve question answering and graph construction tasks.  Battagalia et al \cite{peterb} provide a rich disquisition on graph neural network architectures discussing variety of design features incorporating message passing neural network, non-local neural networks a further variety of graph network variants.
\subsection{GRAPH EMBEDDING APPROACHES}
 Embedding graphs into a low dimensional space encompass a set of techniques that deal with the transformation of an input graph into its respective vector representation mapping it to a point in space with a function mapping. 
 A variety of graph embedding techniques find application in visualization, community detection, localization of wireless devices, power grid network routing, for example. Graph embedding techniques focus on preserving the proximity such that vertices within the same neighbourhood share nearby euclidean space. Historically, graph embedding methods have been been successfully used for obtaining low-level graph data representations. Consider the ISOMAP \cite{tenenbaum} in which a neighbourhood ball is used to convert the data points into a graph, using Djikstara's algorithm to compute the geodesic distance between the vertices. Another approach, the Multidimensional Scaling (MDS) \cite{mds} which when applied to the matrix of geodesic distances results in a reconstructed manifold and is often used to locate well-formed manifolds of complex datasets. Locally Linear Embedding\cite{saul}, considered a variant of PCA, reduces the dimension of the data using a nearest neighbour approach. \cite{jureembed} provide a brilliant treatise on the use of auto-encoders to generate graph representations. Following their notation, we have a \emph{pairwise} encoder-decoder framework to get a pair of embeddings $(z_{i},z_{j})$ such that on reconstruction we have the following loss function $\mathcal{L}$.
\begin{equation}
\begin{split}
    DEC(ENC(v_{i},v_{j})) = DEC(z_{i},z_{j}) \sim S_{\mathcal{G}}(v_{i},v_{j}) 
    \\
    \\
    \mathcal{L} = \sum_{(v_{i},v_{j}) \in \mathcal{D}} l(DEC(z_{i},z_{j}), S_{\mathcal{G}}(v_{i},v_{j}))
    \end{split}
\end{equation}
The general idea is that the encoder \emph{ENC} maps the vertices to vector embeddings and the decoder \emph{DEC} accepts a set of embeddings and decodes the user-specified graph statistics from these embeddings. The general setup  adopted by a majority of authors is to find a similarity function defined on the graph, followed by a pairwise encoder-decoder that learns the embedding and $\mathcal{L}$ is a loss function which determines the performance. A number of methods combine neural learning techniques with natural language are discussed in the following sections. 
\\
\\
\noindent The use of \textbf{Natural Language Techniques} as a tool  \cite{mik1},\cite{mik2},\cite{mik} for learning vertex and edge representations of graphs and has dominated early research in  learning graph representations. The intuition to encode word tokens into a vector in some $\mathcal{N}$ dimensional space where each dimension would represent some semantic meaning in speech for example, is adapted with a bold conjecture for the case of learning graph representations in a similar manner such that each vector would encode the topological information of the graph. In the following paragraph, we discuss some natural language processing theory basics.
\\
\\
\textbf{Notion of context}: Differentiating between the notion of \emph{vertex context} and \emph{word context}, 
we define  \emph{word context}  as  set of  words surrounding the  current word  to be of length $l_{c}$ in a given corpus of $n$ words formed with a window of size $k$ around word $w$. Sliding windows result in dynamic word contexts. We define a  \emph{vertex context} as a set of those vertices composed into a set through the process of some graph traversal algorithm. A vertex context  will automatically encode the topological information around the vertex neighbourhood that is particularly determined by the choice of some graph algorithm.  
\\
\\
\textbf{Probabilistic language models}: A natural start is to use a probabilistic language model defined for using the word to be predicted, from a set of $n$ words as $target \, t$, given $target \, history \, h$. The target $ t$ is defined as the probability of the next word given the target  history  $h$ which is defined as a set of preceding words. The training set is typically a sequence of words $w_{1} \ldots w_{T}$. The model is as follows.
\begin{equation}
p(w_{1}\ldots w_{T}) = \prod_{t=1}^{T} p(w_{}\,|\underbrace{ w_{t-1},\ldots,w_{(t-n+1)}}_{\text{\emph{h}}})
\end{equation}
Continuous bag-of-words (CBOW) model uses \emph{context} $c$, a set of words, from which a word is to be predicted or generated. 
\begin{equation}
p(w_{t} \,| c) = \frac{exp(v_{w_{t}}^{T} \cdot v'_{w_{t}})}
{\sum^{v}_{w=1}exp(v_{w_{t}}^{T}\cdot v'_{w_{t}})} 
\label{eq:26}
\end{equation}
Here, $v_{w}$ corresponds to the input vector representation of word $w$ and $v'_{w_{t}}$ represents the output vector representation.
Skip-Gram, proposed by Mikolov et al, maximises the probability of the surrounding words in context $c$ given the current word $w$. Following notation from \cite{levy}, the conditional probability using the softmax is as follows. Here, $v_{c}$ and $v_{w}$ $\in$ $\Re^{d}$ are vectors representing $c,w$ respectively, $C(w)$ is the set of contexts of word $w$ and $\mathcal{D}$ is the set of all word and context pairs.
\begin{equation}
p(c|w; \theta) = \frac{exp(v_{c}\cdot v_{w})}{\sum_{c'\in \mathcal{C}}exp(v_{c^{'}} \cdot v_{w})}
\label{eq:27}
\end{equation}
Using optimal parameters, $\theta$, we maximise the following objective. The intuition here is that similar words have similar vecctors.
\begin{equation}
\begin{split}
\underset{\theta}{argmax}\prod_{w,c \in \mathcal{C} (w)}   p(c | w; \theta)  = 
\\
\sum_{(w,c) \in D} \Big( log\,exp({v_{c}\cdot v_{w}}) - log \sum_{c^{'}}\,exp({v_{c'}\cdot v_{w}})\Big)
\end{split}
\label{eq:28}
\end{equation}
The idea is to set the parameters in such a manner so that 
$(\ref{eq:28})$ is maximised. However, maximising $(\ref{eq:28})$ may turn out to be expensive and therefore an alternative \emph{Hierarchical Softmax} seems a suitable option.
Proposed by Morin and Bengio \cite{morin}, given a vocabulary $\mathcal{V}$ of $n$ words for a given word $w$ in $\mathcal{V}$, computing the softmax probability of the word would require normalising over the probabilities of all words. Instead of using a flat layer, a hierarchical layer is used to decompose the probabilities of observing the next word in the sequence. The vocabulary of words is converted into a sequence tree structure which is balanced. Now, the probability for a given word is computed using path  to the word from the root followed by the tree path. Modelling probabilities in this manner makes it a cost efficient way of defining a distribution. Negative sampling, an alternative form of Skip-Gram is often used as an efficient way of deriving embeddings but optimises a different objective.
\subsubsection{Recent Advances}
\textbf{Walk based approaches}: Deepwalk \cite{deepwalk} one of the foremost methods to combine deep learning techniques with natural language models to learn representation of random walks on a graph. Deepwalk  uses truncated random walk to transform the sampled linear sequence vertices into a co-occurrence matrix. Skip-Gram model is used to obtain low-dimensional representations for vertices. 
Using social network data, Deepwalk learns social representations that are latent features of the vertices to capture neighbourhood similarity and community membership. It separates the label space from the graph structure to build the feature space and takes an unsupervised approach to capture the network topology using the hierarchical softmax function. 
Deepwalk is suited to work on single large unweighted graphs instead of multiple graphs with a focus on learning similarities between vertices. Nevertheless, it has emerged as one of the most popular baseline models against which new approaches discussed below are measured and developed. 
\\
\\
\cite{Node2vec} is similar to Deepwalk in the manner that it uses random walk (\ref{eq:rwneat}) to generate sequences and that it is scalable to large graphs. It differs in the use of $p$ and $q$ parameters which are preassigned and generate the walks in such a manner that they return to their parent vertex or not far from it. However, several models have to be generated and a subset of labelled vertices are sampled to find the best $p,q$ values. Node2Vec operates in a semi-supervised setting with the graph based objective using stochastic gradient descent. Learning in the network is formulated as a maximum likelihood optimisation problem with two standard assumptions about conditional independence and symmetry in feature space. In a similar spirit, \cite{walk2vec} leverages random walks using two strategies: first to choose initial distributions $\mathcal{P}_{o}$ to produce invariant random walk features $r(\mathcal{P}_{o})$, the second to generate random walk features localised to each vertex. \cite{dnngraph} uses an approach similar to Deepwalk but overcomes the need to use its slow sampling process for generating sequences by using a random surfing model that directly constructs
a probabilistic co-occurrence matrix from a weighted graph. Next, a high- dimensional positive pointwise mutual information matrix (PPMI) is calculated and used as an input to stacked denoising autoencoders to learn the low-dimensional vertex representations. 
\cite{grarep} uses a $k$-step approach, with different $k$ values to capture the relational information amongst vertices from the graph directly. Using global transition matrices defined over the graph overcomes the shortcomings of graph sampling processes that typically involves  tuning parameters such as maximum length of linear sequences and sampling frequency for each vertex. The $k$-step transition probability is used to define a $k$-step loss function over the complete graph. Following Mikolov, noise-contrastive estimation is used to define the objective function. A matrix factorisation approach using singular value decomposition is used to optimise the proposed loss. 
\\
\\
\textbf{Subgraph based embedding approaches}:
In an attempt to overcome the shortcomings of subgraph assumptions (section 4.1.4) in \cite{deepgraphkernels}, 
\cite{subgraph2vec} extends the Weisfeiler-Lehman (section 4.1.2) relabelling strategy defining a radial context to alleviate the problem of selecting fixed-length sequences. The radial Skip-Gram 
captures sequences of varying lengths. The proposed algorithm consists of two main steps: first, the Weisfeiler-Lehman relabelling for building rooted subgraphs; second, the radial Skip-Gram for learning the embeddings of the subgraphs. The motivation for computing the subgraphs is to leverage the \emph{local information} from the neighbourhood of the vertices in order to learn their latent representations. \cite{WLNM} leverages the Weisfeiler-Lehman as a neural machine, that takes the subgraph of a target link, encoding it into an adjacency matrix, for the use case of link prediction. \cite{sub2vec} is another example of learning distributed representations of subgraphs with the goal to embed these into a low-dimensional continuous vector space. Using the local proximity definition that measures how many vertices, edges and paths are shared by two subgraphs, similarity between two subgraphs is learned with with an embedding function. 
In \cite{fastseq}, the authors use a diffusion-like process to extract a subgraph, from which vertex sequences are used to extract features called \emph{hitting frequencies} that are used to construct graph embeddings which are then learned using a neural network. Learned embeddings are used to cluster vertices for the purpose of community detection.
\\
\\
\textbf{Multimodal data graphs}: The authors in \cite{scenegraph} leverage the increasing amount of social media data (section 1.2) to learn joint multimodal embeddings of text and images. A \emph{scene graph}\cite{krishna}, defined by its objects, attributes and relationships is a directed graph constructed as $\mathcal{G} = (\mathcal{O},\mathcal{E})$ where $o \in \mathcal{O}$ is an object in image $\mathcal{I}$ for a given $t \in \mathcal{E}$, a labeled directed edge.  The proposed method learns joint representations of  scene graph  $g$ and image $x$ with respective embedding functions $f_{i}(x)$ and $f_{g}(g)$ that provide continuous representations in $\Re^{D}$ for input images and scene. Three embedding strategies: the bag-of-words, subpath representation and a graph neural network are used to learn the embeddings. In a similar spirit, for the use case of visual question answering, \cite{teney} build graphs over
the scene objects and over the question words. A deep neural network model is used such that one leverages the inherent graph structure in
these representations.
\\
\\
\textbf{Inductive framework}: In \cite{planetoid} the authors propose an \emph{inductive framework}, the case where predictions are made on instances unobserved in the graph at the training time, here the embeddings are defined as a parameterised function of input feature vectors. The model is formulated using a feed-forward neural network, with input feature vector $x$, hidden layer defined as $h^{k}(x)$ = ReLU $(W^{k}h^{k-1}(x) + b^{k})$. The loss function is defined as $\mathcal{L}_{s}$ + $\lambda \mathcal{L}_{u}$. In the transductive formulation $k$ layers are applied on the input feature vector to obtain $h^{k}(x)$ and $l$ layers on embedding $e$ to obtain $h^{l}(e)$ followed by a probability of $p(y \, | x, e)$ of predicting the label $y$ and a transductive loss function. For the inductive case, the label $y$ depends only on the feature $x$ resulting in a respective loss function. Both models are trained using stochastic gradient descent. Another case of inductive learning, \cite{graphsage} proposes a function that generates embeddings by sampling vertex features from its context for unseen data. 
\subsection{PROBABILISTIC APPROACHES}
Probabilistic approaches to learn representations of graph data encompass a variety of neural generative models, gradient based optimisation methods and neural inference techniques. Latent variable modelling involves modelling the relationship between a latent $z$ and an observed variable $x$ with associated parameters $\theta$. 
\begin{equation}
p_{\theta}(x,z) = p(z)p_{\theta}(x|z)
\end{equation}
Here $p_{\theta}(x,z)$ is the joint distribution, $p(z)$ is the prior distribution and $p(x|z)$ is the likelihood. Inference in bayesian models is performed by conditioning on the data calculating the posterior $p(z|x)$. In many cases, calculating the posterior is not straightforward due to complex densities and hence the need for approximate inference tools. The variational approach \cite{blei} encompasses  a general family of methods for approximating complicated densities by a simple class of densities represented by a new approximate posterior distribution $q_{\theta}(z|x)$. A new family of variational methods called  Variational Autoencoders \cite{kingma}  leverages the formulation of neural network using backpropagation techniques to form a new class of neural generative models. The action happens in mapping the data from the latent space to the observed space and that mapping is done using the neural network. 
\begin{figure}[h]
\centering
\begin{tikzpicture}
[line/.style={->}]
\node[obs]                               (x) {$x$};
\node[latent, above=of x ] (z) {$\mathbf{z}$};  
\node[const, above=of z, left= 1.5cm] (theta) {$\mathbf{\theta}$};
\node[const, above=of z, right = 1.5cm] (phi) {$\mathbf{\phi}$};
\edge {z} {x} ; %
\edge {theta}{z};
\edge[dashed]{phi}{z};
\edge {theta}{x};
\draw [blue,line, dashed](x) to [bend left] node {} (z);

\plate {zx} {(x)(z)} {$N$} ;
\end{tikzpicture}
\caption{Graphical Model for the Encoder-Decoder Latent Variable Model}
\end{figure}
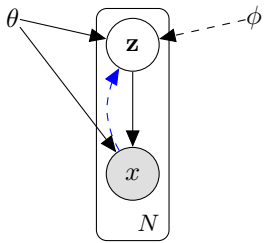
\subsubsection{Recent Advances} 
\noindent \cite{gvae} uses a variational auto-encoding approach to learn the graph representations. A GCN (section 5.3) is used as an encoder and \emph{inner product} as the decoder. The proposed model inference is as follows. 
    \begin{equation}
    q(Z | X, A) = \prod_{i=1}^{N}q(z_{i} | X, A)
    \end{equation}
    $X$ is the vertex features matrix derived using Weisfelier-Lehman (section 4.1.2)  and $A$ is the adjacency matrix with 
    \begin{equation}
    q(z_{i} | X, A)= \mathcal{N} (z_{i} | \mu_{i}, diag(\sigma_{i}))
    \end{equation}
    Here, $\mu$ and $\sigma$ are parameterised by the GCN$(X,A)$. The generative model is given in the form of 
    \begin{equation}
    p(A | Z) = \prod_{i=1}^{N}\prod_{j=1}^{N} p(A_{ij} |z_{i},z_{j}) = \sigma(z_{i}^{T}z_{j})
    \end{equation}
    where  $\sigma(\cdot)$ is the logistic function. Learning is performed by optimising the variational lower bound as follows.
    \begin{equation}
    \mathcal{L} = E_{q}(Z| X, A)[log \, p(A | Z)] - \mathcal{KL}_{q}[(Z| X, A) || p(Z)]
    \end{equation} 
\textbf{Application to Molecular Data}: In \cite{bomb}, the focus is on
generating discrete string representations of chemical molecules using a VAE. Discrete chemical molecules are represented using SMILES (Simplified Molecular Input Line Entry System) format . The encoder network takes each molecule and converts it into a vector into latent space (the space of all such vectors is termed the latent space). The decoder then reproduces a corresponding SMILES string. The network is trained using gradient based optimisation. However, the decoder could associate high probability to strings that are not valid SMILE strings. To address this problem \cite{grammar} introduce a \emph{Grammar Variational Auto-encoder} (GVAE) where a parse tree from context free grammar is used to describe a valid discrete object. The advantage of generating parse trees over text is to ensure that all outputs are valid and follow grammar rules, though their approach does not guarantee chemical validity. \cite{denovo} propose a graph-based generative model for \emph{de novo} molecular design. The authors compare the model's performance against the SMILES based approach by comparing the the rate of valid outputs.
\\
\\
The use of \textbf{Graphical models} for feature space design is championed in \cite{structure2vec}.
A Markov random field is postulated such that a latent variable
$h_i$ is associated with each observed variable $x_i$ according
to a Markovian structure motivated by a corresponding graph
data point. Representations are motivated by the posterior distribution
of each latent variable given observations, which are implemented
by a set of equilibrium equations mapping features of the posterior
of the neighbours of a vertex to its own posterior features, as inspired
by the equations of loopy belief propagation and other variational methods.
Such distributional features are tweaked by supervised learning within a given 
prediction problem, resulting in a representation of the graph data point.
\\
\\
\textbf{Generating Graphs} using a neural generative setting is an open problem. Real-world graphs, with edge connections between vertices are formed through arbitrary connections. Generating graphs involves discrete decisions which may not be differentiable and \cite{smallgraphs} addresses this issue by using a decoder to return a fully-connected graph of a predefined size. A neural network is devised to translate vectors in a continuous space to graphs with the output matched accordingly with a graph matching algorithm.
In \cite{graphgenerative} deep generative graph models, that make no structural assumptions, use a deep neural network to learn the distributions over any arbitrary graphs. A \emph{graph net}, transforms the graph into a sequence of actions, with modules that provide the probabilities such as $(f_{addnode},f_{addedge})$, adding nodes and edges respectively, to help build the structure of the graph offering the opportunity to use a number of different generative models. Scope remains in terms of node ordering, reduction in the sequence of decisions, scalability and overcoming the challenges in training the neural network.
\section{FUTURE DIRECTIONS}
\textbf{Some of the emerging research} in the field looks at the problem of encoding graph data within the prior distribution, learning representations of weighted graphs, learning representations of temporal graphs, learning representations of temporal motifs, addressing non-euclidean graph domain specific challenges, addressing challenges of working with directed graphs. 
There remains further scope in developing novel probabilistic methods to learn representations of graph data.
In this paper, five main approaches: Kernel approaches, Convolution approaches, Graph Neural Network approaches, Graph embedding approaches and Probabilistic approaches have been identified, grouped, surveyed and discussed. Whilst representation learning tackles the leveraging of information from image, sound and text data as an automated process, there does not exist a universally preferred method for working with graph data. A practitioner could use this review as a road map to  gain knowledge about the recent advances and  also as a tool to guide further experimentation.
\section*{Acknowledgment}
Mital Kinderkhedia is thankful to Engineering and Physical Sciences Research Council UK (grant number EP/G036306/1) for funding the doctoral research programme. Many thanks to Christopher Jefferson, Andrew Elliott, Matt Kushner, Brooks Paige and Jon Crowcroft for recommendations and insightful discussions. 
\bibliographystyle{unsrt}  


\end{document}